\documentclass[10pt,twocolumn,times,twoside]{article}

\usepackage{framed,multirow}
\usepackage{multicol}

\usepackage{amssymb}
\usepackage{amsmath} 
\usepackage{adjustbox}
\usepackage{latexsym}
\usepackage{array}
\usepackage{arydshln}
\usepackage{multirow} 
\usepackage{multicol} 
\usepackage{diagbox}
\usepackage{tikz}
\usepackage[font=small,labelfont=bf]{caption}

\usepackage{authblk} 
\usepackage{url}
\usepackage{xcolor}
\definecolor{Prune}{RGB}{99,0,60}

\usepackage[hyperref=true,
			natbib=true,
			backref=true,
			backend=biber,
			url=false,%
        	doi=false,%
        	isbn=false,%
			minbibnames=5,
			maxbibnames=10,
			maxcitenames=1,
			mincitenames=1,
			maxalphanames=1,
			style=authoryear,
			sorting=nyt]{biblatex}
\usepackage[pdftex,pdfborder={0 0 0},
			colorlinks=true,
			linkcolor=Prune,
			citecolor=Prune,
			linktocpage=true,
			]{hyperref}

\usepackage{blindtext}

\newenvironment{figureth}{%
		\begin{figure}[h!]
			\centering
	}{
		\end{figure}
		}

\newenvironment{tableth*}{%
		\begin{table*}[h!]
			\centering
	}{
		\end{table*}
		}
		
\newcommand{\Lsim}{\mathcal{L}_{sim}}
\newcommand{\Lseg}{\mathcal{L}_{seg}}
\newcommand{\Lreg}{\mathcal{L}_{reg}}
\newcommand{\Lsmooth}{\mathcal{L}_{smooth}}	
\newcommand{\Ljac}{\mathcal{L}_{jac}}
\newcommand{\Linv}{\mathcal{L}_{inv}}
\newcommand{\Jdet}{|J_\Phi|}

\newcommand\derivation[2]{\dfrac{\partial #1}{\partial #2}}

\usepackage{fancyhdr}

\newdimen\headlineindent             
\def\abstract
   {%
   \centerline{\normalsize\bf Abstract}%
   \vspace*{12pt}%
   \it%
   }

\newtoggle{cvprrebuttal}
\newtoggle{cvprfinal}
\toggletrue{cvprfinal}
\togglefalse{cvprrebuttal}

\makeatletter

\def\@maketitle
   {
   \newpage
   \vspace*{-1.5cm}
      {\Large \bf \@title \par}
      \vspace*{24pt}
      \normalsize
      \lineskip .5em
        \@author  
      \par
      \vskip .5em
      \vspace*{12pt}
   }

\renewcommand\section{\@startsection {section}{1}{\z@}%
           {18\p@ \@plus 6\p@ \@minus 3\p@}%
           {9\p@ \@plus 6\p@ \@minus 3\p@}%
           {\normalsize\bfseries\boldmath}}
\renewcommand\subsection{\@startsection{subsection}{2}{\z@}%
           {12\p@ \@plus 6\p@ \@minus 3\p@}%
           {3\p@ \@plus 6\p@ \@minus 3\p@}%
           {\normalfont\normalsize\itshape}}
\renewcommand\subsubsection{\@startsection{subsubsection}{3}{\z@}%
           {12\p@ \@plus 6\p@ \@minus 3\p@}%
           {\p@}%
           {\normalfont\normalsize\itshape}}

\makeatother
                                    
\title{MICS : Multi-steps, Inverse Consistency and Symmetric deep learning registration network}%

\author[1,2]{Théo Estienne\thanks{Corresponding author:\\ \texttt{theo.estienne@centralesupelec.fr}}\thanks{Now at Deemea}}
\author[2]{Maria Vakalopoulou}
\author[1,2]{Enzo Battistella}
\author[1,2]{Theophraste Henry}
\author[1,2]{Marvin Lerousseau}
\author[2,3]{Amaury Leroy}
\author[3]{Nikos Paragios}
\author[2]{Eric Deutsch}

\affil[1]{Université Paris-Saclay, CentraleSupélec, Mathématiques et Informatique pour la Complexité et les Systèmes, Inria Saclay, 91190, Gif-sur-Yvette, France.}
\affil[2]{Université Paris-Saclay, Institut Gustave Roussy, Inserm, Radiothérapie Moléculaire et Innovation Thérapeutique, 94800, Villejuif, France.}
\affil[3]{Therapanacea, Paris, France}

\date{}

\begin{document}
\maketitle

\begin{abstract}

Deformable registration consists of finding the best dense correspondence between two different images. Many algorithms have been published, but the clinical application was made difficult by the high calculation time needed to solve the optimisation problem. Deep learning overtook this limitation by taking advantage of GPU calculation and the learning process. However, many deep learning methods do not take into account desirable properties respected by classical algorithms. 

In this paper, we present MICS, a novel deep learning algorithm for medical imaging registration. As registration is an ill-posed problem, we focused our algorithm on the respect of different properties: inverse consistency, symmetry and orientation conservation. We also combined our algorithm with a multi-step strategy to refine and improve the deformation grid. While many approaches applied registration to brain MRI, we explored a more challenging body localisation: abdominal CT. Finally, we evaluated our method on a dataset used during the Learn2Reg challenge, allowing a fair comparison with published methods.
\end{abstract}




\section{Introduction}

Deformable registration is studied for many years in the medical imaging field. It consists of finding the best correspondences between two volumes. In traditional approaches (i.e. non-deep learning), registration was formulated as an optimisation problem with one similarity term and one regularisation term (see \citep{sotiras2013deformable}). Different algorithms were proposed such as graph-based \citep{glocker2008dense,heinrich2013mrfbased}, flows of diffeomorphism \citep{beg2005computing,yancao2005large}, symmetric normalization \citep{avants2008symmetric} or stationary velocity field based \citep{arsigny2006logeuclidean,ashburner2007fast}. One limitation of classical methods relies on the fact that the optimisation problem must be resolved for each new pair of patients, preventing real-time application. The development of deep learning enables real-time for two main reasons. The calculation time is reduced thanks to GPU implementation, and the optimisation problem is resolved only during the training phase and not for the prediction of new pairs. 

One significant difficulty of registration is the evaluation of its performance. Registration is an ill-posed problem as many different transformations could warp one image to another, even using affine transformation. Evaluating the performance in terms of organs or landmarks correspondences is one of the standard ways to benchmark the performance of the registration problem. Such an evaluation does not reflect the quality of the deformation grid. To overcome this problem and to measure the grid's plausibility, different metrics have been proposed, such as the standard deviation of the Jacobian \citep{adriandalca2020learn2reg}. 

In machine learning and deep learning, the benchmark of algorithms' performances has often been performed through the organisation of challenges and evaluation on publicly available datasets. Indeed, the problem of overfitting makes it essential to compare algorithms with identical training and testing datasets. Concerning registration, different challenges have been performed on brain MRI \citep{klein2009evaluation}, on thoracic CT during the EMPIRE10 challenge \citep{murphy2011evaluation} and for abdominal CT  \citep{xu2016evaluation}. Recently Learn2Reg, a registration challenge, was organised during the Miccai conference \citep{adriandalca2020learn2reg}. It comprises four different tasks: multimodal interoperative brain registration (CT-US), CT lung inspiration-expiration registration, abdominal CT registration and MRI hippocampus registration. Among the best solutions were proposed multi-scale deep learning approach \citep{mok2020large}, probabilistic dense displacements network \citep{heinrich2019closing}, a classical approach using Markov Random Field \citep{heinrich2013mrfbased} and a deep learning approach based on pretraining and spatial gradients \citep{estienne2021deep}. In this paper, we performed our evaluation on the third task of the Learn2Reg Challenge. Evaluating our approach on a public dataset allows a fair comparison with our proposed method and other algorithms.

Recently, different reviews detail deep learning-based registration, its development, the different approaches and the future trends \citep{haskins2020deep,boveiri2020medical,fu2020deep}. More precisely, \citet{boveiri2020medical} analysed the most studied organs in registration's papers. Brain MRI is far ahead of other organs, with more than 70 papers when they wrote their survey. The brain has several advantages: the absence of large deformation due to the skull's surface or the existence of large public datasets. In this article, we explored the application of deep learning registration to abdominal CT. This task is also very challenging difficult because of the large deformation in the abdomen and the lack of correspondence. 

This article propose a deep learning algorithm that learns to register abdominal CTs while respecting suitable topological properties. Our convolutional network project both images to a common latent space and merge them to output the deformation gradient. To raise the registration performance, we investigate a multi-step strategy and the use of pretraining using many public abdominal datasets. The pretraining is also upgraded by the introduction of pseudo-segmentations produced by a segmentation network. The pseudo-segmentations allow weakly supervised pretraining while the public datasets collected do not have segmentation masks. Concerning the smoothness of the deformation grid, we study the impact of two new regularisation losses, and we introduce symmetry constraints as well as a multi-step formulation to refine and identify small deformations. The training is split into two parts, the pretraining with supplementary data and pseudo segmentations and the fine-tuning using only the challenge's data. Our main contributions are the following : 
\begin{enumerate}
\item Optimising our network with three different regularisation losses to respect symmetry, inverse consistency, local orientation conservation;
\item Developing a multi-step framework to refine the predicted deformation taking advantage of our symmetric formulation.
\item Taking advantage of publicly available data to pretrain and fine-tune our network. 
\end{enumerate}
 
In the section \ref{section:related}, we present existing research on deep learning registration and regularisation strategies. Section \ref{section:method} describes our methodology, including our symmetric approach and the multi-step formulation. Finally, section \ref{section:experiments} presents our experiments performed on abdominal CT and the results.
\section{Related work}
\label{section:related}

Deep learning-based registration is often separated into three different categories: supervised, unsupervised and weakly supervised. Other classifications can be found into \cite{fu2020deep,haskins2020deep,boveiri2020medical}. The term supervision refers to the use of ground truths deformation to train the network. These ground truths grids were generated by classic algorithms or by artificial deformations. Currently, unsupervised and weakly supervised formulations are more explored as they do not need pre-calculated transformations. Instead, they use the spatial transformer  \citep{jaderberg2016spatial} to warp the moving image in the deep learning framework and calculate the error at the image level. Unsupervised approaches have been developed for different organs such as brain MRI \citep{balakrishnan2018unsupervised,dalca2018unsupervised}, cardiac MRI \citep{krebs2018unsupervised,devos2019deep} or multimodal approach \citep{ferrante2018adaptability,qin2019unsupervised}. Weakly supervised methods augment unsupervised ones by using supplementary labels, for instance, segmentation masks or biological landmarks. The labels are not passed as input of the networks. Instead, the deformation grid warped them, and a supplementary loss is calculated between the warped labels and the target labels. Thanks to this framework, the network only needs labels during the training and the prediction of a new transformation is made only using imaging data. Such methods are developed in \citep{hu2018weaklysupervised,hering2018enhancing,balakrishnan2019voxelmorph,estienne2019uresnet}.

Various modifications have been to proposed deep learning-based formulations. We will focus on those who bring more regularisation on the predicted transformations. In classical registration algorithms, different regularisation techniques have been explored, for instance, flows of diffeomorphisms with LDDMM framework \citep{beg2005computing,yancao2005large} or symmetric normalisation formulation \citep{avants2008symmetric}. Another regularisation strategy is to respect several properties, including symmetry, inverse consistency, or diffeomorphism \citep{sotiras2013deformable}. In deep learning approaches, constraints are often created by the addition of supplementary losses or by modifications to the network architecture. 

Among the constraints, different researchers explored methods to regularise the determinant of the Jacobian matrix. Negative values of the Jacobian translate local non topological behaviour of the deformation, while it also provides us with valuable information about the local conservation of the orientation. Thus, enforcing the Jacobian positivity will help to generate a more realistic deformation. Different formulation of Jacobian constraints can be found in \citet{mok2020fast, kuang2019faim, zhang2020diffeomorphic}. \citet{hering2020constraining} proposed a different approach by controlling the changes of the volume. Another formulation was developed by \citet{mansilla2020learning} to take into account anatomical constraints. They added a supplementary denoising autoencoder to the registration pipeline, and they constrained the encoding of the warped segmentation to be as close as possible to the encoding of the target segmentation. Applying the loss at the encoding scale will create a global matching witch is not produced by pixel level loss.

Recent formulations introduced symmetry properties through CycleGAN approaches. In this framework, two networks are needed. The first one maps the volumes from the $X$ space to the $Y$, while the other does the inverse. Two cycle-consistency losses are then introduced,  compelling the composition of the two networks to return to the original image. CycleGAN have produced excellent results for image-to-image translation \citep{zhu2017unpaired}. \citet{kim2019unsupervised} introduced the cycle formulation to ensure topological preservation and tested it for registration of the liver anatomy. CycleGAN formulations are well suited for multi-modality registration, where the moving and target images are in different spaces. This was proposed by \citet{wang2019fire} to perform T1-weighted to T2 Flair brain MRI registration. Finally, \citep{wang2020ltnet} combined cycle formulation and weakly supervised registration, using two cycle-losses, one calculated on images and the other one on segmentation masks.

Multi-scale or multilevel formulations are another strategy to smooth the transformation and reduce local perturbations. UNet architecture is widely spread among unsupervised registration methods. It could be considered as a multilevel formulation as down-sampling and up-sampling operations are applied during the forward pass. However, in the registration context, we consider a formulation as multi-scale if deformations at different resolutions are explicitly calculated and applied to the source image. Such approaches are proposed in  \citep{mok2020large,hering2019mlvirnet,fechter2020oneshot}. They differ by the way the different level transformations are combined and the training strategy.

Finally, multi-step formulations refine predicted deformations by applying the registration network between the deformed image and the fixed image. These approaches are also known as cascade networks, recursive registration networks or multi-stage networks. Such formulations were applied to the liver and the heart and proposed in \citep{zhao2019recursive,zhao2020unsupervised,devos2019deep}. Differences between these approaches lie in the use of one or several independent networks or in the training procedure, which can be performed in an end-to-end way or one network at a time and then freezing it before passing to the next one.

\citep{mok2020fast} is the closest method of to our study. They proposed a deep learning diffeomorphic and symmetric approach. Their diffeomorphic formulation is based on stationary velocity field and scaling and squaring procedures. Thus, our method has some significant differences, including weakly supervision, the symmetric formulation, the inverse consistency loss and the multi-step formulation.

\section{Method}
\label{section:method}

\subsection{Registration}
\label{section:registration}

Let consider two volumes $A$ and $B$ defined over a $n$ dimensional space $\Omega$. For the rest of this paper, $n$ will be equal to $3$. In most of the registration algorithms, the volumes are denominated as $F$ and $M$ for respectively fixed and moving volumes. However, we chose this designation as $A$, and $B$ have a symmetrical role in our approach. The registration goal is to find the best deformation $\Phi$ to warp one image into another. Our network $g_{\theta}$ takes as input $A$ and $B$ and return the two transformations $\Phi_{A\rightarrow B}$ and $\Phi_{B\rightarrow A}$. Two main formulations exist in deep learning-based deformation. In the first one, the network outputs the displacement field $u$ and the deformation field $\Phi$ is obtained by $\Phi = Id + u$. This formulation has the advantage of simplicity. For small displacements, the inverse transformation can be approximated as $\Phi^{-1} \approx Id - \textbf{u}$. However this approximation is invalid with large deformation \citep{ashburner2007fast}. The second formulation, called \emph{diffeomorphic} \citep{dalca2018unsupervised,krebs2018unsupervised,mok2020fast}, model the deformation as the flow of a diffeomorphism. The network outputs $v$, a stationary velocity field (SVF), and the deformation $\Phi$ is obtained by {scaling and squaring} algorithm \citep{ashburner2007fast,arsigny2006logeuclidean}.  In this paper, we used a third approach based on gradients. The network predicts the gradient along axis $x$, $y$ and $z$ of the deformation $\Phi$. Then we obtained $\Phi_x$ (resp, $\Phi_y$, and $\Phi_z$) by integration of $\nabla_x \Phi_x$ (resp, $\nabla_y \Phi_y$, and $\nabla_z \Phi_z$). A cumulative sum operation approximates the integration operation. This formulation helps us to keep the order of warped voxels. Indeed, if we constraint the gradient to be positive, we guarantee that two neighbouring pixels will stay in order. However, as we consider only the gradients along one axis, we do not have insurance to produce a diffeomorphism. Indeed we only put a positive constraint on the diagonal of the Jacobian matrix and not on its determinant. For more details, the interested reader should consult \citet{stergios2018linear,shu2018deforming,estienne2020deep}. After integrating, we obtained the warped image using spatial transformer \citep{jaderberg2016spatial}.

One major difference between other deep learning methods and ours is how volumes are passed through the network. Most of the methods in the litterature concatenate the two volumes before passing them through the network \citep{balakrishnan2019voxelmorph,devos2019deep,krebs2019learning}.  The network process then a four dimension volume including $x$, $y$ and $z$ axis and two channels representing each volume. Lets consider that our network $g_\theta$ is divided into one encoder $E$ and one decoder $D$. Following \citet{estienne2020deep,estienne2021deep}, we passed the volumes $A$ and $B$ independently through the encoder $E$. Then we merged the encoding of $A$ and $B$ through the subtraction operation, and we obtained the gradients by applying the decoder to it. Finally, the expressions of our spatial gradients and our warped volumes are : 
\begin{equation}
\label{eq:main}
\begin{split}
	\nabla \Phi_{A\rightarrow B} &= D(E(A) - E(B)) \\
	\nabla \Phi_{B\rightarrow A} &= D(E(B) - E(A)) \\
	A_{warp} &= \mathcal{W}(A,\Phi_{A\rightarrow B}) \\
	B_{warp} &= \mathcal{W}(B,\Phi_{B\rightarrow A})
\end{split}
\end{equation}
where $\mathcal{W}$ denotes the backward trilinear interpolation sampling and $A_{warp}$ and $B_{warp}$ the two warped volumes.

This formulation has several advantages. First, the subtraction operation respects suitable mathematical properties. Indeed, if we calculated the registration between two identical volumes, the subtraction will give a zero tensor which will then generate identity transformation. Furthermore, the opposite of the transformation is obtained by inverting $A$ and $B$ in the subtraction. Secondly, our formulation need only one network to generate the forward and backward transformation, while CycleGAN methods use one network for the forward transformation and one network for the backward \citep{kim2019unsupervised,wang2020ltnet}. In this framework, the encoder can be seen as playing the role of a feature extractor, as the two volumes are processed independently.

\subsection{Network architecture}

Our network is based on the 3D UNet architecture \citep{cicek20163d}. It consists of a symmetrical encoder-decoder architecture with four blocks with respectively 64, 128, 256 and 512 channels, resulting in 20 million parameters. Each block includes $3$D convolution layers with kernels of size $3$, instance normalisation layer, leaky ReLU activation function. The up and down-sampling operations are performed by $3$D convolution with kernel size $2$ and stride $2$, and the encoder and decoder are connected through skip connection. The final layer predicts a three channels tensor between $0$ and $1$ thanks to the sigmoid operation, which is integrated through cumulative sum to obtain the transformation. The warped image and segmentation are then produced using the spatial transformer. 

\subsection{Multi-steps formulation}

To improve the registration produced by our network, we implemented a multi-steps strategy. The idea is to refine the deformation by predicting the transformation between the deformed image and the target. This strategy can be applied either during the training of the network or during the inference. Lets define the initial grid as the grid at step one : $\nabla \Phi_{A\rightarrow  B}^{1} = D\left( E(A) - E(B)\right)$. The warped image $A$ at step one is then : $A_{warp}^{1} = \mathcal{W}\left(A, \Phi_{A\rightarrow B}^{1}\right)$. We can now define recursively the deformation grid and the deformed image at step $i$ : 

\begin{equation}
\begin{split}
\nabla \Phi_{A\rightarrow B}^{i} &= D\left( E(A_{warp}^{i-1}) - E(B)\right) \\
A_{warp}^{i} &= \mathcal{W}\left( A_{warp}^{i-1}, \Phi_{A\rightarrow B}^{i}\right)
\end{split}
\label{eq:multisteps}
\end{equation}

The multi-step deformation of the image $B$ is obtained by inverting $A$ and $B$ in the equation \ref{eq:multisteps}. Our multi-step formulation takes advantage of our strategy to combine the two volumes (eq \ref{eq:main}). Indeed, we only need to pass the deformed image at step $i$ through the encoder and then merge it with the target's encoding. Thus, we gained calculation time and memory usage. 

\subsection{Pretraining and Noisy segmentations}

Transfer learning is a common strategy to improve the performance of deep learning algorithms. In two dimensions, people often initialise networks with weights coming from the ImageNet dataset. However, in three dimensions, there is no consensus about pretraining strategies.  Thus recent publications proposed three dimensions pretrained networks with tasks such as segmentation \citep{chen2019med3d} or unsupervised pretraining \citep{zhou2019models}.

To improve transfer learning performance, we proposed a pretraining strategy using directly the registration task instead of using a different task like segmentation or reconstruction. Our proposed pipeline is: First, train the registration task with a large dataset built from public medical databases and then fine-tune the network with only the official dataset. One advantage of registration compared to other tasks is that the network can be trained in an unsupervised way. Thus we can only collect data and do not need to have corresponding labels. However, weakly supervised registration has proved to outperform unsupervised training \citep{hu2018weaklysupervised,hering2018enhancing,balakrishnan2019voxelmorph}. For this reason, we experiment with the influence of weakly supervised pretraining. 

As the supplementary data do not have the same segmentation mask as our original dataset, we have to design a strategy to obtain these labels. One possibility would have been to generate the segmentation masks by clinicians, who would have also been required to devote an excessive amount of time to perform annotations. Instead, we decided to train a segmentation network to generate the segmentation labels automatically without human interaction. As these labels have been acquired by an automatic and trained algorithm, we designate them as \emph{pseudo labels}. Our segmentation network is a 3D UNet \cite{cicek20163d} trained with the following parameters : batch size equal to 6, learning rate equal to $1e^{-4}$, leaky ReLU activation functions, instance normalisation layers and random crop of patch of size $144\times144\times144$. Depending on the dataset, each image has different organs segmented by doctors. Thus we used a modified Dice loss to back-propagate only the available labels. We apply various post-processing steps to improve the pseudo labels' quality: keep the ground-truths labels for the organs available, keep only the biggest connected component of the predicted label to remove small segmentation and manual inspection of the predicted segmentation to remove outlier results.  

Our pipeline is defined as follows: \textit{i)} Train a segmentation network using available masks \textit{ii)} Predict the pseudo labels using the trained network \textit{iii)} Train a new network on the registration task, using supplementary data and pseudo segmentations \textit{iv)} Fine-tune the registration network using only the challenge data for the task in question and ground-truths segmentations.

\subsection{Loss Function}

We train our network by minimising a combination of five different losses. Theses loss will have two different goals: obtaining the best transformation to warp $A$ to $B$ and ensuring that the grid respect desirable topological properties. Moreover, as we developed a symmetrical approach to the registration problem, each loss will also have a symmetrical formulation. 

The first two losses, the similarity loss $\Lsim$ and the segmentation loss $\Lseg$, constraint the network to produce the best deformation. The similarity loss is the mean square error between the deformed image and the target image. Other approaches substitute the mean square error by the local cross-correlation for $\Lsim$ \citep{balakrishnan2019voxelmorph,mok2020fast,mansilla2020learning}, however, for our problem local cross-correlation did not really help during the training. The segmentation loss consists of a Dice loss \citep{milletari2016vnet} between the ground-truth segmentation and the deformed segmentation and thus is a supervised loss contrary to $\Lsim$. Many recent articles show that weakly-segmentation registration outperforms unsupervised registration \citep{hu2018weaklysupervised,hering2018enhancing,balakrishnan2019voxelmorph}. The network can then learn to produce deformation not only in function of image intensity but also from organs. Their expression is the following :  

\begin{align}
\Lsim &= ||A_{warp} - B ||^2 + ||B_{warp} - A ||^2 \\
\Lseg &= \text{Dice}\left(A_{warp}^{seg}, B^{seg}\right) + \text{Dice}\left(B_{warp}^{seg}, A^{seg}\right)
\end{align}

We added three supplementary losses to force our network to produce realistic deformations. The regularisation loss $\Lreg$ control the smoothness of the predicted deformation. The Jacobian loss $\Ljac$ impose the positivity of the Jacobian and the inverse consistency loss $\Linv$ force the predicted grids to be the inverse of each other.

The regularisation loss (or smooth loss) is one of the most common losses in deep learning-based registration. Indeed, minimising only $\Lsim$ and $\Lseg$ could create non-realistic deformations. This loss penalises the high value of the gradient to enforce the smoothness of the grid. Contrary to most of the approaches, our network predicts the gradient directly (see section \ref{section:registration}). 

However, the smooth loss is not enough to have a grid which respect topological properties. It does not prevent folding and wrong orientation. A new loss was proposed in recent papers \citep{mok2020fast,kuang2019faim,zhang2020diffeomorphic} to impose positive values of the determinant of the Jacobian matrix (also called the Jacobian). The expression of the Jacobian matrix of the deformation $\Phi$ at one voxel $p$ is the following : 
\begin{equation}
J_\Phi(\textbf{p}) = \begin{pmatrix} \derivation{\Phi_x}{x} & \derivation{\Phi_x}{y} & \derivation{\Phi_x}{z}\\
\derivation{\Phi_y}{x} & \derivation{\Phi_y}{y} & \derivation{\Phi_y}{z}\\
\derivation{\Phi_z}{x} & \derivation{\Phi_z}{y} & \derivation{\Phi_z}{z}\\
\end{pmatrix}(\textbf{p})
\end{equation}
We can then calculate the determinant for each voxel $p$ and obtain the Jacobian in the same form as the moving and fixed volumes.  The Jacobian characterise two properties of the local behaviour of the deformation. First, the sign of the Jacobian informs us of the local orientation of the deformation field. If its value at voxel $p$ is negative, the registration will reverse the orientation locally around $p$. On the opposite, it will conserve the orientation if the Jacobian is positive. Secondly, the transformation will be locally invertible around $p$ if the Jacobian is non-zero at $p$. Thus, we want to compel the Jacobian to be strictly positive. The Jacobian loss $\Ljac$ will be the sum of all negative values of the Jacobian.

Our last loss will constraint the network to generate symmetric transformations witch are the inverse to each other. Many recent deep learning formulations do not respect these properties. Our formulation is symmetric by construction, as we predict both $\Phi_{A\rightarrow B}$ and $\Phi_{B\rightarrow A}$. Though, we do not have guarantees that the two transformations are effectively the inverse of each other. Therefore, we implemented a new loss to respect the inverse consistency properties. This loss consists in penalise the difference between the composition of the two transformations and the identity transformation. The composition of the transformation is performed using the spatial transformer. Another formulation of the inverse consistency loss was proposed in \citet{zhang2018inverseconsistent}. 

The mathematical formulation of our three regularisation loss is the following : 
\begin{equation}
\Lreg = ||\nabla \Phi_{A\rightarrow B}|| + ||\nabla \Phi_{B\rightarrow A}||
\end{equation}
\begin{equation}
\begin{split}
\Ljac = \frac{1}{N} \sum_{p \in \Omega}{\max \left(0, - |J_{\Phi_{A\rightarrow B}}|\right) }&\\ 
+ \max \left(0, - |J_{\Phi_{B\rightarrow A}}|\right)&
\end{split}
\end{equation}
\begin{equation}
\begin{split}
 \Linv =   &|| \Phi_{A\rightarrow B} \circ \Phi_{B\rightarrow A} - \Phi_{Id} || \\
 + &|| \Phi_{B\rightarrow A} \circ \Phi_{A\rightarrow B} - \Phi_{Id} ||  
\end{split}
\end{equation}

Finally, our total loss will be the combination of these five different loss with their respective weight $\alpha$, $\beta$, $\gamma$, $\delta$ and $\epsilon$ : 

\begin{equation}
\mathcal{L} = \alpha \Lsim + \beta \Lseg + \gamma \Lreg + \delta \Ljac + \epsilon \Linv
\end{equation}
In the multi-steps formulation, we applied the total loss $\mathcal{L}$ to each of the warped volumes, deformation and grid : $\mathcal{L}_{multi} = \sum_{i} \mathcal{L}(
\nabla \Phi_{A\rightarrow B}^{i}, \nabla \Phi_{B\rightarrow A}^{i} , A_{warp}^{i}, B_{warp}^{i})$. As the registration is an ill-posed problem, the tuning of the weights will have a high impact on our network's performance. A high value for the regularisation weights will produce a grid close to the identity and poor performance in organ registration. In contrast, a too-small value will produce unrealistic deformed volumes. We will explore the contribution of these weights in the following sections.

\section{Experiments}
\label{section:experiments}

\subsection{Training parameters}
\label{section:parameters}

We performed experiments to study the impact of the pseudo segmentations, the different regularisation losses and the multi-step strategy. For each experiment, we first pretrained the network with a big dataset and pseudo segmentations and then fine-tuned with a smaller dataset and the ground-truths segmentations. For all our experiments, we choose the following hyper-parameters: the learning rate was set equal to $1e^{-4}$, the network processed randomly cropped patches of size $128\times 128\times 128$ during the training and $256\times 160 \times 192$ during the prediction. The batch size was equal to $4$ and $2$ for the training with respectively one step and two steps. We applied three HU windows to the CT scans: the abdominal, lung and bones windows. The corresponding width and level values are $W=400$, $L=40$ (abdominal), $W=1400$, $L=-500$ (lung), $W=1000$, $L=400$ (bones). Therefore, the network's input is a three channels volume, and the similarity loss is calculated with these three windows. We did not apply any data augmentation procedures, as it seems to lower the measured performance. Other hyper-parameters such as the values of the weights $\alpha$, $\beta$, $\gamma$, $\delta$ and $\epsilon$ are given in each subsection.

\subsection{Dataset}

To generate the pseudo labels for our registration problem, we built a dataset combining many publicly available datasets. The first dataset comes from the Learn2Reg Challenge \citep{adriandalca2020learn2reg}. This dataset was first used in a comparison of registration algorithms \citep{xu2016evaluation}. It comprises 50 abdominal CT and thirteen abdominal organs segmented: spleen (Spl), right kidney (RKid), left kidney (LKid), gall bladder (GBla), oesophagus (Oes), liver (Liv), stomach (Sto), aorta (Aor), inferior vena cava (InfVe), portal and splenic vein (P\&SVein), pancreas (Pan), left adrenal gland (LAd), and right adrenal gland (RAd). All volumes have different characteristics (pixel spacing, image size, for instance). Thus, organisers applied the following processing steps: affine registration, resampling to 2 mm voxel size, cropping to a common shape of $256\times 192\times 160$. We split this data into training, validation and test set with respectively 20, 10 and 20 patients, using the same split as during the Learn2Reg challenge. We collected three supplementary abdominal datasets : the Medical Segmentation Decathlon or MSD \citep{simpson2019large}, the Kits 2019 dataset \citep{heller2020kits19} and the TCIA Pancreas dataset  \citep{roth2016data,clark2013cancer}.  The MSD challenge was a segmentation challenge where the participants had to segment different structures (organs, tumours, vessels) with different modalities. We selected three sub-datasets of the MSD corresponding to abdominal CT:  Liver (Task 3), Pancreas (Task 7) and Spleen (Task 9) with respectively 200, 420 and 61 volumes.  The Kits 19 dataset contains 300 abdominal CTs with the segmentation of kidney and kidney's tumours. The TCIA Pancreas comprises 82 CT scans with the segmentation of 8 different organs: spleen, left kidney, gallbladder, oesophagus, liver, stomach, pancreas and duodenum. The segmentations have been published in parallel with two articles, first focusing only on the pancreas \citep{roth2015deeporgan}, then on seven supplementary organs \citep{gibson2018automatic}. We kept only the segmentations already available in the L2R dataset. In the case of tumour segmentation, we chose to merge the tumour's labels with the corresponding organs. All these datasets have different imaging characteristics. We standardised them by resampling them to $2$ mm voxels size (similar to the L2R dataset) and affine registration with the software Ants \citep{avants2009advanced}. The affine registration was performed with the same image of the training set of the L2R dataset. A summary of the different volumes used is depicted in Table \ref{tab:dataset}.

\begin{table}
\centering
\begin{adjustbox}{max width=\linewidth}
	\begin{tabular}{|l|cc|}	 \hline
		\multirow{2}{*}{Dataset} & \multirow{2}{*}{Segmentations} & Number \\ 
		 &  & of Volumes \\ \hline
		Learn2Reg & 13 organs & 50 \\
		TCIA Pancreas &  8 organs &  82 \\
		Kits 19  & kidney \& tumour & 300\\
		MSD Liver  & liver \& tumour  & 200\\
		MSD Pancreas & pancreas \& tumour & 420\\
		MSD Spleen & spleen  &61  \\ \hline

	\end{tabular}
\end{adjustbox}
	\caption{An overview of the different dataset used for this study}
	\label{tab:dataset}
\end{table}
\subsection{Metrics}

To evaluate our registration performance, we use different metrics. We need to measure both the accuracy and robustness of the method as well as the smoothness of the deformation. We follow the Learn2Reg organisers using the Dice score, 30\% of the lowest Dice score, Hausdorff distance and standard deviation of the log Jacobian (noted SdLogJ in this chapter). 
For all the experiments, the metrics are calculated with the script given by the Learn2Reg's organisers and on the ten patients of the validation set (45 volume pairs). Using the same metrics and the same validation set, we allow a fair comparison with others results published on this challenge.

\subsection{Impact of the pseudo-segmentations}

\begin{figure*}[h]
  \includegraphics[width=1\linewidth,trim=2.2cm 4.2cm 4cm 3cm,clip=true]{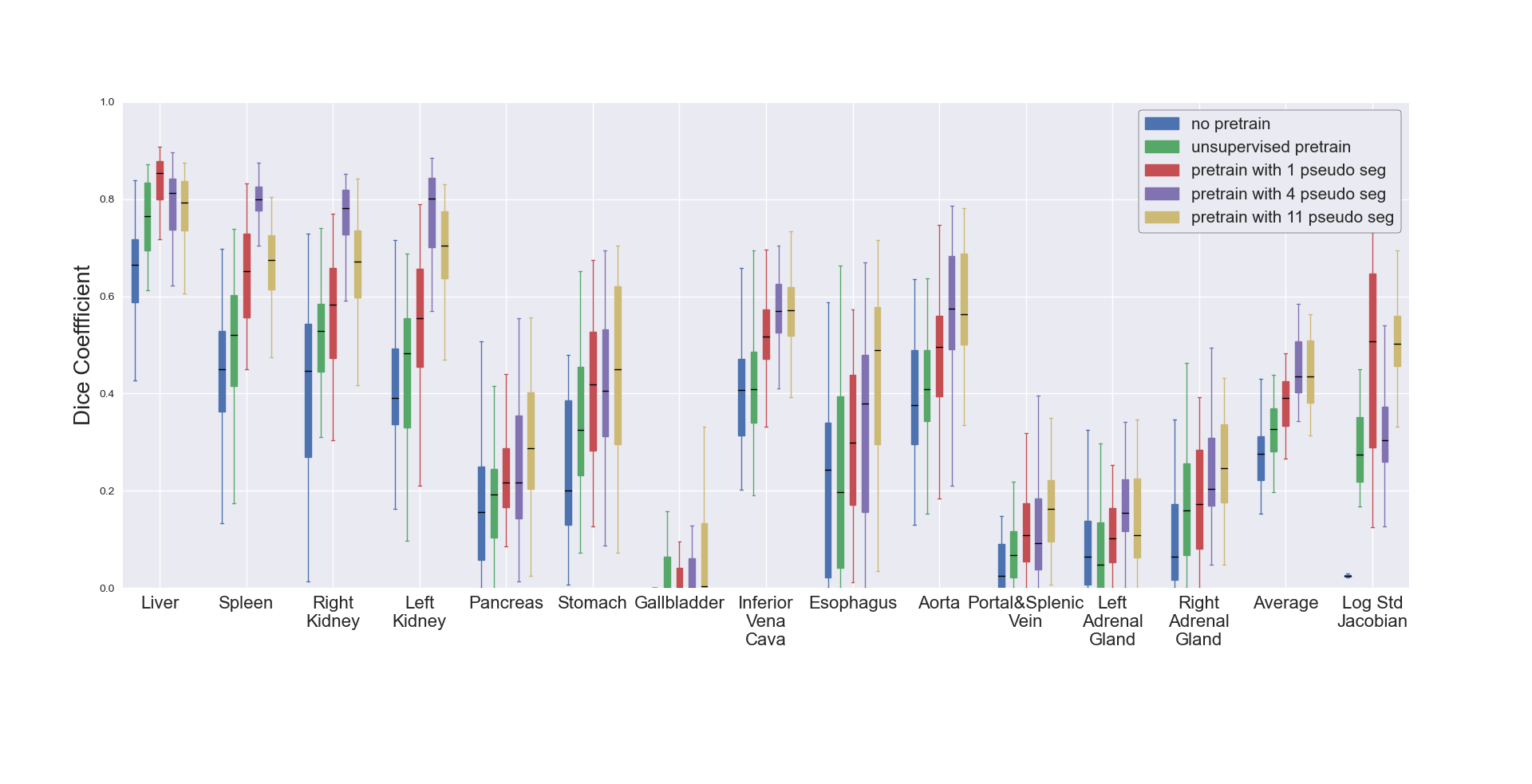}
  \caption[Comparison of the impact of the number of pseudo-segmentations during the pretraining]{Comparison of the impact of the number of pseudo-segmentations during the pretraining. The Dice coefficient is represented in a box plot for the 13 evaluated organs as the average Dice and the Standard Deviation of the Log-Jacobian. The results are displayed for the ten patients included in the Learn2Reg validation set (45 pairs), and five different pretraining strategies are compared.}
  \label{fig:pretrain_impact}
\end{figure*}

We performed experiments to measure the impact of the pseudo-segmentations during the pretraining and present the results in Figure \ref{fig:pretrain_impact}. For all experiments, we kept the same training parameters. The weights of the different losses were set to $\alpha = \beta = \gamma = 1$ and $\delta = \epsilon = 0$, meaning that only the smooth loss $\Lsmooth$ is used to regularise the transformations. The pretraining was performed during $24$ hours (approximatively $100$ epochs) with 760 different volumes, and the network was fine-tuned with 300 epochs using only the 20 patients of the Learn2Reg dataset and the ground-truth segmentations. The training lasted in total $38$ hours, and we selected the epoch with the minimal loss on the validation set. 

We compared five different experiments: the registration network without any pretraining, an unsupervised pretraining without using the pseudo-segmentations, and three supervised pretraining using respectively 1 (Liv), 4 (Liv, Spl, RKid, LKid) and 11 organs as pseudo-segmentations. The experiment with 11 organs correspond to using all the organs as pseudo-segmentations except the right and left adrenal glands as these organs are small, and the segmentation network could not predict them with a very good accuracy. After the training, our segmentation network achieve the following performances in terms of dice : $0.92$ (Spl), $0.90$ (RKid), $0.91$ (LKid), $0.94$ (Liv) $0.83$ (Sto),  $0.74$ (Pan),  $0.72$ (GBla), $0.89$ (Aor), $0.76$ (InfV), $0.62$ (PorV) and $0.61$ (Oes). The segmentation's validation set comprises 21 patients coming from the Learn2Reg and the TCIA Pancreas dataset. The best Dice performances were achieved by the biggest organs such as the liver or spleen, and the organs present in large quantity in the dataset we built.  

In Figure \ref{fig:pretrain_impact}, we represent a box plot of the registration performances for the five experiments in terms of the Dice coefficient. The Dice coefficient is evaluated for the 13 organs, and we also represent its average value and the smoothness of the transformations using the SdLogJ. We can draw several conclusions from this figure. First, unsupervised pretraining improves the performance of the majority of the organs but has a higher impact on voluminous organs such as the liver and spleen. Second, when we use some organs' pseudo segmentation during the pretraining, the performance is increased mainly on these organs but also on others. For instance, the experiment which used only the liver during the pretraining resulted in Dice's improvement not only for the liver but also for the spleen, kidneys or stomach. Adding only a few organs produce a sort of overfitting on these. Indeed, if we compare the experiments with 1, 4 and 11 organs, we find a Dice's decrease in the liver, spleen, and kidneys. The network concentrates more on the supervised organs than on other body parts, and thus the performance decreases when we add new organs. However, the overall dice still improve. Finally, the number and the choice of organs used as pseudo segmentations influence the grid's regularity. The experiment using only the liver and the one with 11 organs produce transformations with low regularity (high SdLogJ). In one case, the network focuses too much on the liver, producing a noisy grid. In the other case, the grid becomes more complex and, thus, more irregular. For all the experiments, it is important to recall that all the organs were used during the fine-tuning step. 

From these experiments, we can conclude that adding organs helps register, even with approximated segmentations. The most voluminous organs have the biggest impact.

\subsection{Impact of the regularisation losses}

\begin{figureth}
  \includegraphics[width=1\linewidth,trim=1.2cm 0.5cm 2.5cm 2cm,clip=true]{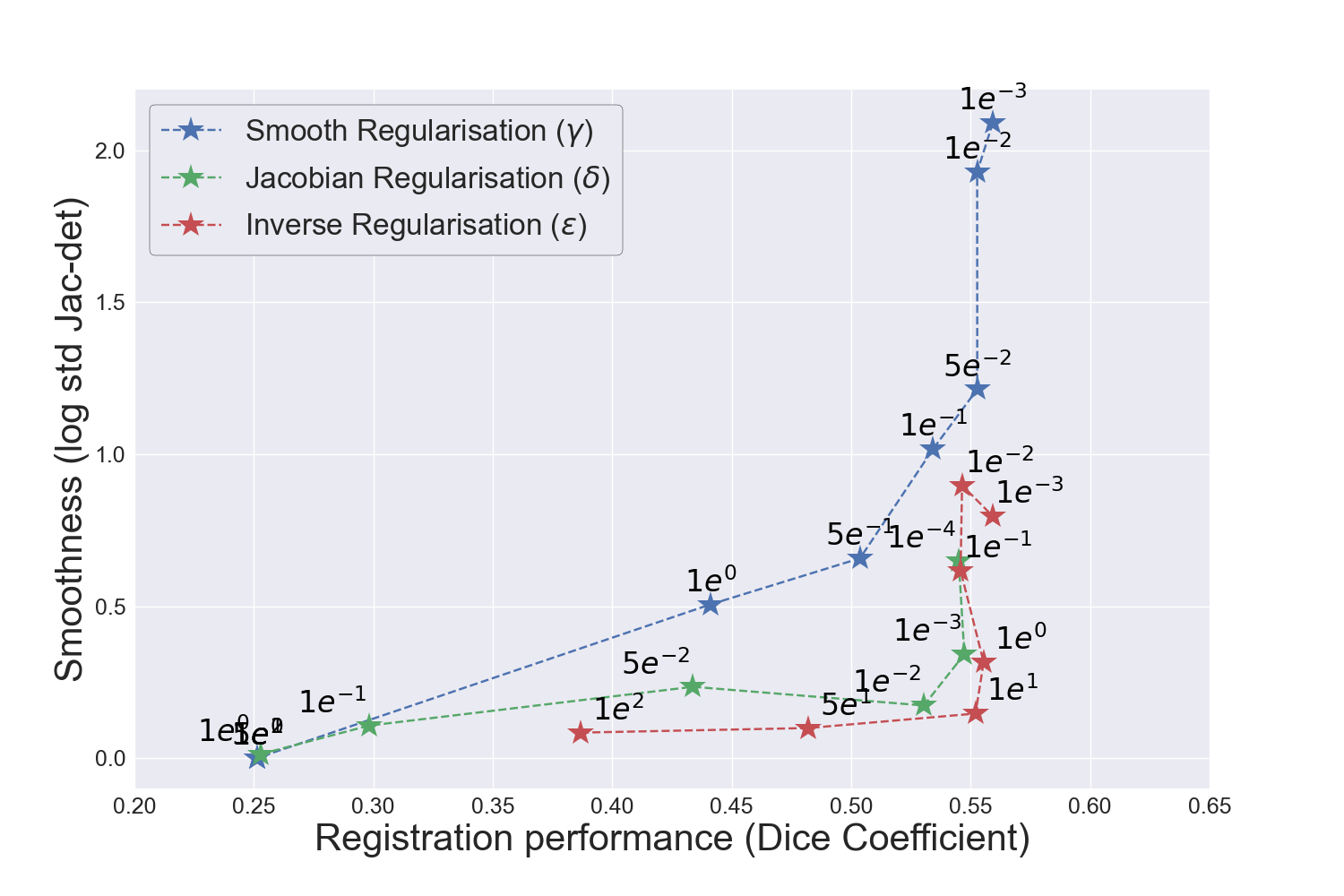}
  \caption{Representation of the smoothness of the grid in function of the registration performances for different regularisations weights. The metrics were calculated for the 10 patients from the Learn2Reg validation set (45 pairs). The value of the weights $\gamma$, $\delta$ and $\epsilon$ are indicated on the graph and correspond to the blue, green and red curve.}
  \label{fig:dice_noise}
\end{figureth}

We explored the impact of the different regularisation losses $\Lsmooth$, $\Ljac$ and $\Linv$. We kept the same training parameters as for the previous section, changing only the values of $\gamma$, $\delta$ and $\epsilon$ : pretraining with pseudo-segmentations of 11 organs for 24 hours and then fine-tuning for 300 epochs. The value of the weights was the same during the pretraining and the fine-tuning. We present the results in the Figure \ref{fig:dice_noise}. This figure aims to study the correlation between registration's performance in terms of Dice and the grid's smoothness with respect to the SdLogJ. The best transformations correspond to a high dice and a low standard deviation, which is the bottom right corner of the graph. The bottom left corner corresponds to the identity transformation (very regular grid and low Dice), and the top right corner to a good registration in Dice but with a poor regularity.  

The blue curve of Figure \ref{fig:dice_noise} represents the performance with $\delta$ and $\epsilon$ equal to $0$ and $\gamma$ varying between $1e^{-3}$ and $1e^{1}$. We showed that there is a strong correlation between the smoothness and performance using only $\Lsmooth$. Especially, low values of the weight $\gamma$ result in very noisy transformations while they do not improve a lot the registration performance in terms of Dice. Thus the smooth loss $\Lsmooth$ is insufficient to obtain a high Dice together with very regular deformations.

For the red and green curve of Figure \ref{fig:dice_noise}, we set $\gamma$ equal to $1e^{-1}$, as this value reaches both good high performance in terms of Dice and smoothness, and we varied the $\delta$ and $\epsilon$ values. For the green one, $\delta$ iterates between $1e^{-4}$ and $1e^0$ while $\epsilon$ was constantly equal to $0$ and inversely for the red curve $\delta$ is equal to $0$ and $\epsilon$ goes from $1e^{-3}$ to $1e^2$. The two regularisations losses have a strong impact on the smoothness of the deformations. They allow decreasing the SdLogJ while keeping the Dice around 0.55. However, when $\delta$ and $\epsilon$ reach high values, the deformation is close to the identity transformations. Moreover, the Jacobian loss has a stronger impact than the inverse consistency loss. Indeed the predicted transformations get closer to identity transformations with $\delta$ around $1e^{-1}$, while we observed the same results for values of $\epsilon$ around $1e^{2}$. We obtained the best compromise between the registration performances and the smoothness of the grid for values of $\delta$ between $1e^{-2}$ and $1e^{-3}$ and values of $\epsilon$ between $1$ and $10$.


\subsection{Impact of the multi-steps strategy}

We also explored the influence of the multi-step strategy on the grid's smoothness and the registration's performance. The multi-step formulation is given in equation \ref{eq:multisteps}, and it consists of finding the transformation between the fixed volume and the already deformed volume.

We investigate first the consequence of applying the multi-step formulation during the training phase or during the inference phase. For these experiments, we set the weights to $\alpha=\beta=1$, $\gamma=1e^{-1}$, $\delta=1e^{-2}$ and $\gamma=1e^{1}$, we pretrain the network during 24 hours using 11 organs as pseudo-segmentations and 760 patients and fine-tune the network with the Learn2Reg training set and the ground-truth segmentations. In Table \ref{tab:number_steps}, we present the results both in Dice and SdLogJ for two networks trained with 1 and 2 steps, while the inference's steps are between 1 and 4. The results for the network trained with 1 step shows that using the multi-step strategy during the inference has a negative impact. Indeed, the Dice coefficient decreases from $0.35$ to $0.30$ while the grid regularity is reduced significantly, with the SdLogJ going from $0.13$ to $0.76$. Concerning the two-steps approach, we obtained different results. First, looking at the first line (1 step inference), we obtain a higher Dice and a similar SdLogJ when the network is trained with two steps than one, with a Dice equal to $0.54$ compared to $0.35$. This demonstrates the impact of the dual-step training, even if the inference is performed with one step. For all these experiments, we kept the same hyper-parameters to have fair comparisons between the different setups (1 or 2 steps). Studying the inference with two steps, it increases the Dice coefficient ($0.54$ to $0.60$) but reduces the regularity ($0.12$ to $0.26$ for the SdLogJ). Finally, performing the inference with 3 and 4 steps do not improve the Dice but increase the noise. This experiment demonstrates that training the network with a 2 step formulation improve the results, even if the inference is performed with 1 step. We also show that applying the inference with more steps than the training does not ameliorate the grid but instead makes it noisier.

\begin{tableth*}
\begin{adjustbox}{max width=\linewidth}
\begin{tabular}{c||cccc}
\hline
\multirow{2}{*}{\backslashbox{Inference}{Training}} & \multicolumn{2}{c}{1 step}                         & \multicolumn{2}{c}{2 steps}   \\ \cdashline{2-5} 
 & \multicolumn{1}{c}{Dice} & \multicolumn{1}{c}{SdLogJ} & \multicolumn{1}{c}{Dice} & \multicolumn{1}{c}{SdLogJ} \\ \hline 
1  steps        & $0.35 \pm 0.078$    &   $0.13 \pm 0.019$    & $0.54 \pm 0.070$    &  $0.12 \pm 0.009$                       \\
2  steps               & $0.33 \pm 0.067$    &   $0.52 \pm 0.116$   & $0.60 \pm 0.059$    &  $0.26 \pm 0.050$                       \\
3  steps               & $0.30 \pm 0.067$    &   $0.76 \pm 0.146$    & $0.61 \pm 0.058$    &  $0.34 \pm 0.062$                     \\
4  steps               &        \textbackslash             &          \textbackslash        & $0.61 \pm 0.057$    &  $0.41 \pm 0.069$                      
\end{tabular}
\end{adjustbox}
  \caption{Study of the impact of the multi-steps formulation during the training or the inference phase. Two models are compared trained with 1 or 2 steps and four inference formulation. The metrics are calculated on the Learn2Reg validation set using 45 pairs. The average dice over 13 organs and the standard deviation of the Log-Jacobian are presented.}
  \label{tab:number_steps}
\end{tableth*}

In a second time, we examine the impact of the 2-step strategy for different regularisations. For these experiments, we set the loss weights to $\alpha=\beta=1$, $\gamma=0.1$ and different choices for $\delta$ and $\epsilon$. These two parameters correspond to the Jacobian and inverse consistency loss $\Ljac$ and $\Linv$. We choose three different combinations corresponding to a strong regularisation ($\delta=5e^{-2}$, $\epsilon=5e^1$), a medium regularisation ($\delta=1e^{-2}$, $\epsilon=1e^1$) and a weak regularisation ($\delta=1e^{-3}$, $\epsilon=1e^0$). These weights were selected using the Figure \ref{fig:dice_noise}. Following the results presented in Table \ref{tab:number_steps}, we select the same number of steps for the training and inference, comparing the 1-step and 2-step formulations. We did not experiment on a bigger number of steps, mainly because of memory limitations. We represent the results in terms of registration's performance with the Dice and grid's smoothness with the SdLogJ in Figure \ref{fig:multisteps}, and we also give the corresponding numerical values in Table \ref{tab:multisteps}. In Figure \ref{fig:multisteps}, we draw an arrow going from the one step's results to the two steps' results. From this figure, we see that the 2-step approach boost the results for all the choice of regularisation parameters. Moreover, it raises the performance on the Dice coefficient while keeping the smoothness relatively low. It is also interesting to notice that the 2-step formulation improved more the Dice coefficient when the regularisation is stronger, and the 1-step method produces deformation very close to the identity transformation. Finally, we demonstrate that the dual-step strategy is an improvement of registration formulation, as it increases the accuracy while preserving the smoothness.


\begin{tableth*}
\begin{adjustbox}{max width=\linewidth}
\begin{tabular}{cc||cccc}
\hline
\multicolumn{2}{c||}{Weights}  & \multicolumn{2}{c}{1 Step} & \multicolumn{2}{c}{2 Steps} \\ \cdashline{3-6} 
$\delta$        & $\epsilon$       & Dice              & SdLogJ              & Dice              & SdLogJ          \\  \hline
$1e^{-3}$         & $1e^{0}$       & $0.57 \pm 0.066$  & $0.25 \pm 0.098$       & $0.62 \pm 0.055$  & $0.38 \pm 0.091$   \\
$1e^{-2}$         & $1e^{1}$       & $0.34 \pm 0.078$  & $0.13 \pm 0.019$       & $0.60 \pm 0.059$  & $0.26 \pm 0.050$   \\
$5e^{-2}$         & $5e^{1}$       & $0.29 \pm 0.070$  & $0.066 \pm 0.002$      & $0.44 \pm 0.076$  & $0.080 \pm 0.004$                      
\end{tabular}
\end{adjustbox}
  \caption{Comparison of the 1 step and 2 steps strategy for different values of the regularisation weights. The metrics are calculated on the Learn2Reg validation set using 45 pairs. The Dice coefficient and the SdLogJ are given. For each regularisation weights, the 2 steps strategy improve the Dice coefficient.}
  \label{tab:multisteps}
\end{tableth*}

\begin{figureth}
  \includegraphics[width=1\linewidth,trim=1.2cm 0.5cm 2.5cm 2cm,clip=true]{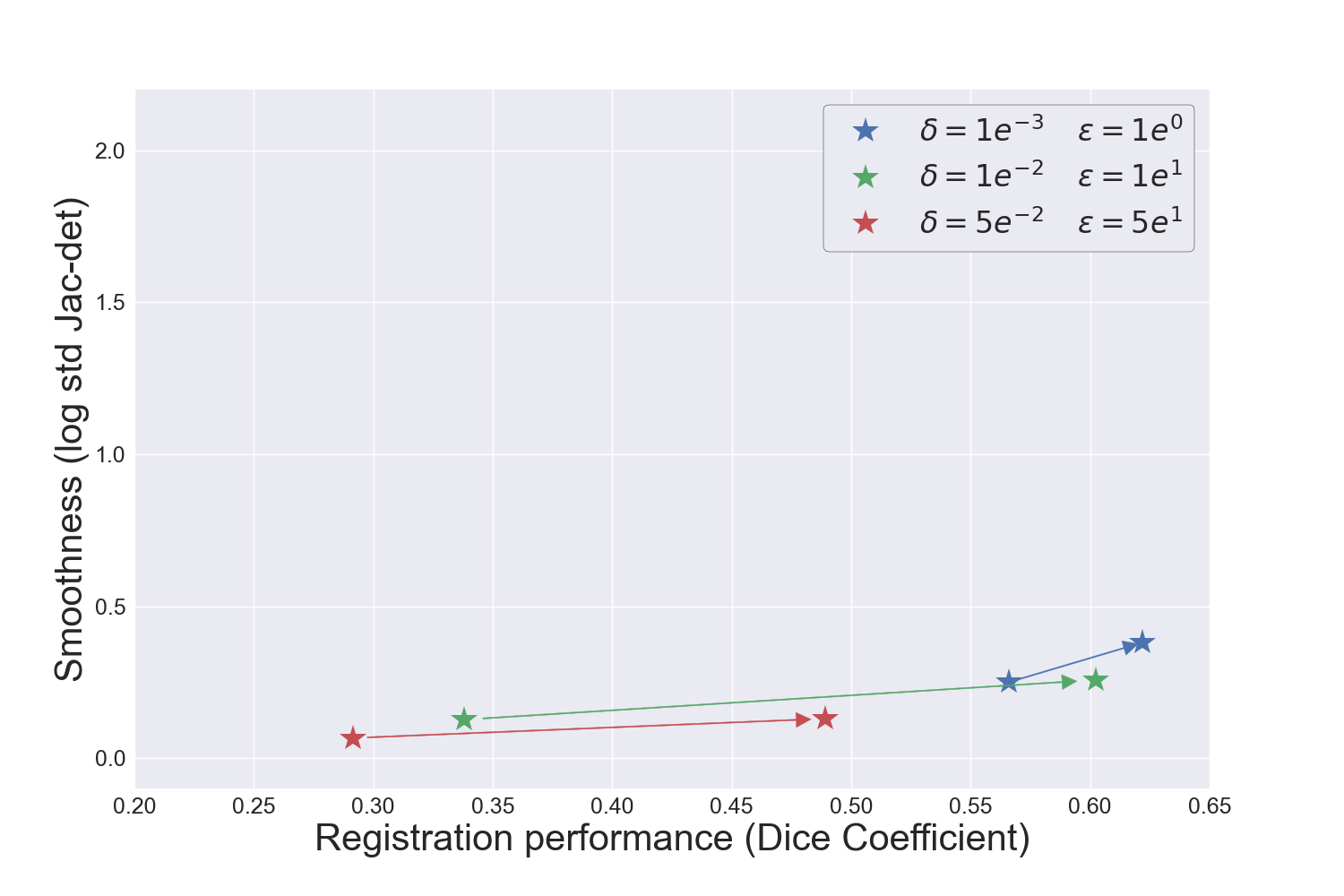}
  \caption{Impact of the multi-step on the smoothness and the Dice coefficient. The arrows go from the results of the 1 step to the results of the 2 steps. Three different training have been made with different values of the regularisation weights $\delta$ and $\epsilon$. The metrics were calculated for the 10 patients from the Learn2Reg validation set (45 pairs). The X and Y axis have the same limits as in Figure \ref{fig:dice_noise}}
  \label{fig:multisteps}
\end{figureth}

\subsection{Comparison with Learn2Reg Results}

Finally, we compare our approach with the results of the Learn2Reg Challenge 2020 and present the results in Table \ref{tab:res_task3}. For our approach, we set the regularisation weights to: $\alpha=\beta=1$, $\gamma=1e^{-1}$, $\delta=1e^{-2}$ and $\epsilon=1e^{1}$ and use our proposed dual-steps strategy. We choose these regularisations weights as they provide a good trade-off between the smoothness and the registration. We pretrain the network during 48 hours using 11 organs as pseudo-segmentations and 760 patients and fine-tune the network with the Learn2Reg training set and the ground-truth segmentations. Other training parameters are given in Section \ref{section:parameters}.

\begin{tableth*}
\begin{adjustbox}{max width=\linewidth}
\begin{tabular}{l||cccc}
\hline
Methods& Dice & Dice30 & Hd95 & SdLogJ \\ \hline
\citep{estienne2021deep} & 0.64 & 0.40 & 37.1 & 1.53 \\
\citep{mok2020large}   & \textbf{0.67} & 0.48  & \textbf{36.5} & 0.12 \\
\citep{heinrich2013mrfbased} & 0.51 & 0.29 & 39.8 & \textbf{0.11}  \\
\citep{heinrich2020highly} & 0.46 & 0.22 & 42.1 & 0.43 \\
MICS (ours) & 0.66 & \textbf{0.61} & 38.8 & 0.21  
  \end{tabular}

\end{adjustbox}
\caption{Comparison of our method (MICS) with the top-performing methods of Task 3 of Learn2Reg Challenge. The metrics are calculated on the test set of the Learn2Reg 2020 Challenge. The first three rows are results obtained by different teams during the challenge. Bold indicates best values.}

\label{tab:res_task3}
\end{tableth*}

\begin{figure*}[th!]

\foreach \X in {Moving, Deformed Volume, Deformed Mask, Deformation Grid, Jacobian}   
{\begin{minipage}[b]{.20\linewidth}
  \centering
  \X
  \end{minipage}%
}


\foreach \moving in {0004,0007,0010}{%
\foreach \imgname in {moving_,deformed_,deformed_mask_,integrated_grid_,Jacobian_}{%
\begin{minipage}[t]{.20\linewidth}
\centering
\includegraphics[width=1\linewidth,trim=6.9cm 2.6cm 6.9cm 2.6cm,clip=true]{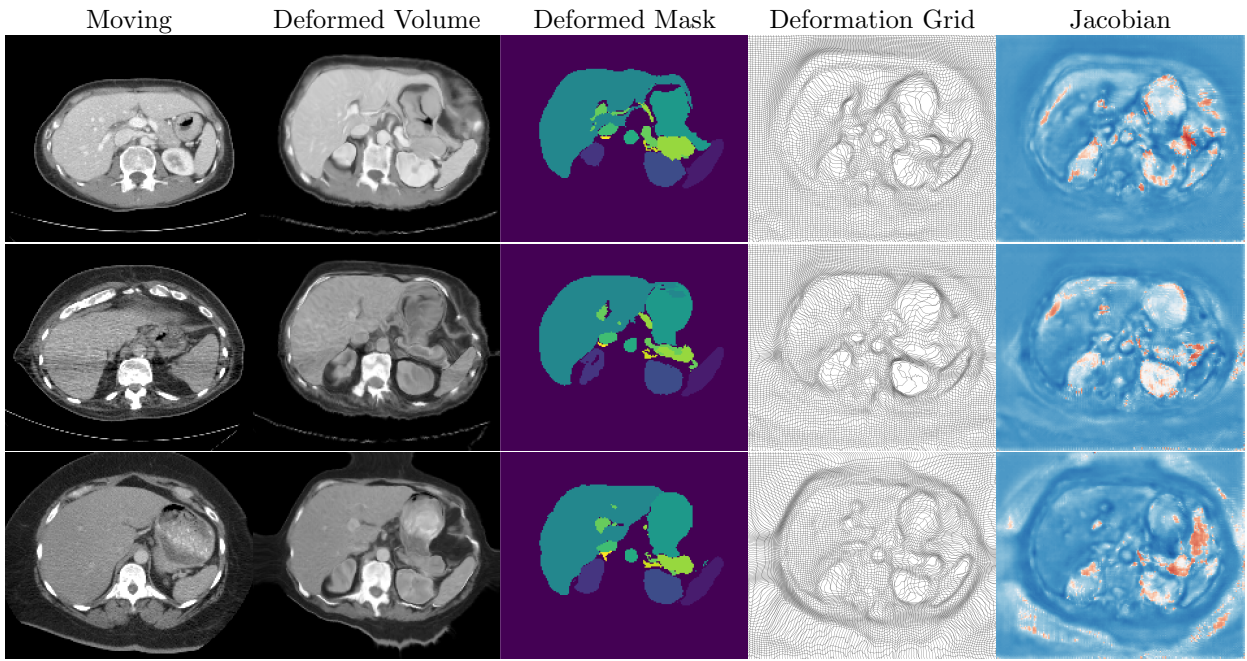}
\end{minipage}%
}

}
  
  \caption{Representation of the registration results for three different moving volumes and the same fixed volume. From left to right: the moving CT, the deformed CT, the deformed segmentation, the transformation $\Phi$ and the Jacobian $\Jdet$. The Jacobian is depicted in blue, white and red for respective positive, zero and negative values.}
  \label{fig:deformed}
\end{figure*}

Our method outperforms three teams of the challenge, obtaining better Dice and Dice30 \citep{heinrich2013mrfbased,heinrich2020highly,estienne2021deep}. Concerning the smoothness of the grid, measured by the SdLogJ, we obtain satisfactory results, overpassing two competitors. Especially if we compare with the results of \citet{estienne2021deep}, which have a close result for the Dice, our SdLogJ is seven times smaller ($1.53$ vs $0.21$), showing the impact of our supplementary regularisation losses. Two methods obtained more plausible deformations, one deep learning-based \citep{mok2020large} and one iterative method \citep{heinrich2013mrfbased}.

Our method reports very promising results, particularly for the Dice30, with a score of $0.61$, while the best method had $0.48$. This attests the robustness of our method, which performed well even for the most challenging moving-fixed volumes pairs. The method proposed by \citet{mok2020large}, based on deep learning and a multi-scale pyramid approach, surpass our results with similar results for Dice and Hausdorff distance but mostly a SdLogJ twice lower. Multi-scales formulations seem to be very efficient to generate accurate and smooth deformation, especially for localisation with high displacements such as abdominal CT. These methods will become state-of-the-art for deep learning-based registration.

In Figure \ref{fig:deformed}, we display the results of MICS for the three different pairs, keeping the same fixed volume and changing the moving volume. We represent the moving CT, the deformed volume and segmentation, the deformation grid and the Jacobian. Even if we trained our network using a specific loss to reduce negative Jacobian, we have negative values (in red in the figure) as well as crossing and folding in the grid.


\section{Discussion and Conclusions}
\label{section:conclusion}

In this paper, we present MICS a deep learning-based registration method. Our network predicts spatial gradient which are integrated to obtain deformation grid. We take a special attention to the respect of registration properties, including symmetry, inverse-consistency and local non topological deformations. We study, in particular, the impact on the accuracy and the plausibility of the deformations of three different regularisation losses, the smooth loss, the Jacobian loss and the inverse consistency loss. In addition, we provide an extensive study of the effect and impact of the pseudo labels. Our results are improved by two other techniques : a dual-step formulation and a supervised pretraining using pseudo segmentations. Finally, we benchmarked our new formulation with the Learn2Reg 2020 Challenge dataset, comparing it with the top-performing results of the challenge and our initial submission. We demonstrate that both the use of regularisation losses and the multi-step formulation improve the performances of the registration approach.

We also highlighted the difficulty of evaluating registration algorithms. How should we rank two algorithms, one being irregular but accurate, the other being smooth but less accurate? This issue is of particular importance in the case of deep learning-based registration, where networks manage to elaborate complex but noisy deformations. During the Learn2Reg Challenge \citep{adriandalca2020learn2reg}, the organisers set various weights for each metric (calculation time, accuracy, smoothness, robustness), but altering these weights could lead to a new winner solution. The representation of both the smoothness and the accuracy in one graph allows a new visual evaluation and a better comparison of different methods. We provided such representations in Figure \ref{fig:dice_noise} and \ref{fig:multisteps}.

One important limitation of our work is the presence of negative Jacobian, despite the Jacobian loss. A stronger regularisation could remove them. However, it would produce a deformation close to the identity transformations. In the same way, the composition of the forward and backward transformations lead to a grid unequal to the identity transformations. The deformed images are finally irregular and not applicable in a clinical application, demonstrating the requirement of great improvement for abdominal registration. In addition to the multi-step formulation described in this chapter, multi-scale approaches obtained promising results for different localisation \citep{hering2019mlvirnet,fechter2020oneshot,mok2020large}. These approaches produce accurate and smooth deformations thanks to the combination of transformations at different resolutions levels.

In the future, we want to explore a clinical application of our method. More specifically, we would like to apply our network to multi-temporal follow-up of patients, monitoring disease progression. Temporal registration of the same patient should produce less noisy deformations than inter-patient registration. One application of abdominal registration could be the case of liver tumour, to monitor the tumour growth inside the liver. The major difficulty would be to disjoin the deformations due to the tumour growth from the one related to natural deformations (digestion or breathing, for instance). Finally, we want to investigate the predictive power of registration networks for other clinical tasks such as survival prediction in cancerology. Three methodologies could be examined: i) using the registration network as a pretraining step for a network that predicts clinical parameters, ii) exploring the correlation between the deformation grid predicted by registration networks and clinical features and iii) using the latent space which encode the deformation grid and connecting a simple classifier to it.

\printbibliography

\end{document}